\documentclass{llncs}
\usepackage[german,english]{babel}
\usepackage{times,theorem}
\usepackage{latexsym,color,graphics}
\usepackage{diagrams}
\diagramstyle[tight,centredisplay%
,dpi=600%              for more modern laser-printers
,UglyObsolete%         formerly called ``noPostScript''
,heads=LaTeX%
]
\usepackage{epsfig}
\usepackage{amssymb}
\usepackage{amsmath}
\usepackage{amsfonts}
\usepackage{xspace}
\usepackage{graphicx}
\begingroup
\catcode`\~=11
\gdef\urltilde{\lower 0.6ex\hbox{~}}
\endgroup

\newcommand{\A}{\mathcal{A}} 
 \newcommand{\D}{\mathcal{D}}
\newcommand{\E}{\mathcal{E}} \newcommand{\F}{\mathcal{F}}
 
\newcommand{\I}{\mathcal{I}} 
 \renewcommand{\L}{\mathcal{L}}
\newcommand{\M}{\mathcal{M}} \newcommand{\N}{\mathcal{N}}
 \renewcommand{\P}{\mathcal{P}}
 \newcommand{\R}{\mathcal{R}}
\renewcommand{\S}{\mathcal{S}} \newcommand{\T}{\mathcal{T}}
 \newcommand{\V}{\mathcal{V}}
\newcommand{\W}{\mathcal{W}}

\title{Intensional FOL: Many-Sorted Extension}
\author{Zoran Majki\'c}
\authorrunning{Zoran Majki\'c}
\institute{ISRST, Tallahassee, FL, USA\\
\email{majk.1234@yahoo.com}
}

\newtheorem{propo}{Proposition}
\newtheorem{coro}{Corollary}

\begin{document}

\maketitle
\begin{abstract}
  The concepts used in IFOL have associated to them a list of sorted attributes, and  the sorts are the intensional concepts as well.   The requirement to extend the unsorted IFOL (Intensional FOL) to many-sorted IFOL is mainly based on the fact that a natural language is implicitly many-sorted and that we intend to use IFOL to support applications that use natural languages. Thus, the proposed version of many-sorted IFOL is just the completion of this conceptual feature of the IFOL.
 \end{abstract}

\section{Introduction to Unsorted Intensional FOL}
  Contemporary use of the term "intension" derives from the
traditional logical doctrine that an idea has both an extension and
an intension. Although there is divergence in formulation, it is
accepted that the extension of an idea  consists of the subjects to
which the idea applies, and the intension consists of the attributes
implied by the idea. In contemporary philosophy, it is linguistic
expressions (here it is a logic formula), rather than concepts, that
are said to have intensions and extensions. The intension is the
concept expressed by an expression of intensional algebra $\A_{int}$, and the extension is the set of items to which the expression applies. This usage resembles use of Frege's use  of "Bedeutung" and "Sinn" \cite{Freg92}.

Intensional entities (or  concepts) are such things as Propositions,
Relations and Properties (PRP). What make them "intensional" is that they
violate the principle of extensionality; the principle that
extensional equivalence implies identity. All (or most) of these
intensional entities have been classified at one time or another as
kinds of Universals \cite{Beal93}.  We begin with the informal theory that universals (properties (unary relations),
relations, and propositions in PRP theory \cite{Beal79}) \index{PRP theory} are genuine entities that bear fundamental logical relations to one another.
\begin{definition} \label{def:Idomain} \textsc{Intensional logic PRP domain $\D$:}\\
In intensional logic the concepts (properties, relations and
propositions) are denotations for open and closed logic sentences,
thus elements of the structured domain   $~\D = D_{-1} + D_I$, (here
$+$ is a disjoint union) where
\begin{itemize}
  \item A subdomain $D_{-1}$ is made of
 particulars (individuals).
  \item The rest $D_I = D_0 +
 D_1 ...+ D_n ...$ is made of
 universals (\emph{concepts})\footnote{In what follows we will define also a language of concepts with intensional connectives defined as operators of the intensional algebra $\A_{int}$ in \cite{Majk22}, so that $D_I$ is the set of terms of this intensional algebra. }: $D_0$ for  propositions with a distinct concept $Truth \in D_0$, $D_1$ for properties
 (unary concepts)  and  $D_n, n \geq 2,$ for n-ary concept.
\end{itemize}
\end{definition}
  The  concepts in $\D_I$ are denoted by $u,v,...$, while the
 values (individuals) in $D_{-1}$ by $a,b,...$ The empty tuple $<>$ of the nullary relation $r_\emptyset$ (i.e. the unique tuple of 0-ary relation) is an individual in  $D_{-1}$, with $\D^0 =_{def} \{<>\}$. Thus, we have that
  $\{f,t\} = \P(\D^0) \subseteq \P(D_{-1})$, where by $f$ and $t$
 we denote  the empty set $\emptyset$ and set $\{<>\}$ respectively.

  Let $~\mathfrak{R} = \bigcup_{k \in \mathbb{N}} \P(\D^k) = \sum_{k\in \mathbb{N}}\P(D^k)$ be the set of all k-ary relations, where $k \in \mathbb{N} =
\{0,1,2,...\}$. Notice that $\{f,t\} = \P(\D^0) \subseteq \mathfrak{R}$,
that is, $f,t \in\mathfrak{R}$ and hence the truth values are extensions in $\mathfrak{R}$.

 The \emph{intensional interpretation} is a mapping between the set $\L$ of formulae of the logic language  and  intensional entities in $\D$, $I:\L \rightarrow \D$, is a kind of  "conceptualization", such that  an open-sentence (virtual predicate)
 $\phi(x_1,...,x_k)$ with a tuple of all free variables $(x_1,...,x_k)$ is mapped into a k-ary \emph{concept}, that is, an intensional entity  $u = I(\phi(x_1,...,x_k)) \in D_k$, and (closed) sentence $\psi$ into a proposition (i.e., \emph{logic} concept) $v =  I(\psi) \in D_0$ with $I(\top) = Truth \in D_0$ for the FOL tautology $\top \in \L$  (the falsity in the FOL is a logic formula $\neg\top \in \L$).

 A FOL language constant (nullary function) $c$ is mapped into a particular $~ I(c) =_{def} I_T(c)  \in D_{-1}$ if it is a proper name, otherwise in a correspondent concept in $\D$. Thus, in any application of intensional FOL, this intensional interpretation that determines the meaning (sense) of the knowledge expressed by logic formulae is \emph{uniquely determined (prefixed)}.
 However, the extensions of the concepts (with this prefixed meaning) vary from a context (possible world, expressed by an extensionalizzation function) to  another context in a similar way as for different Tarski's interpretations of the FOL:
\begin{definition} \label{def:extent} \textsc{Extensions and extensionalization functions:}\\
 The extensions of the intensional entities (concepts) are given by the set
 $\E$ of extensionalization functions $h:\D \rightarrow
 D_{-1} +\mathfrak{R}$, such that
 \begin{equation}\label{eq:ExtenFunct}
 h = h_{-1} + h_0 + \sum_{i\geq 1}h_i:\sum_{i
\geq -1}D_i \longrightarrow D_{-1} + \{f,t\} + \sum_{i\geq
1}\P(D^i)
\end{equation}
 where $h_{-1}:D_{-1} \rightarrow D_{-1}$ for the particulars, while
$~h_0:D_0 \rightarrow \{f,t\} = \P(\D^0)$ assigns the truth values
in $ \{f,t\}$ to all propositions with the constant assignment
$h_0(Truth) = t = \{<>\}$, and for each $i\geq 1$, $h_i:D_i \rightarrow \P(D^i)$
assigns a relation to each concept.

Consequently, intensions can be seen as \verb"names" used in natural languages of atomic or
composite concepts, while the extensions correspond to various rules that
these concepts play in different 'explicit' worlds $h$. We denote by $\hbar$ the actual explicit world and by $\E$ a given set of '\emph{explicit}' worlds.
\end{definition}
From a logic point of view, two possible worlds $w$ and $w'$ are
indistinguishable if all sentences have the same extensions in them,
so that we can consider an extensionalization function $h$ as a
"possible world", similarly to the semantics of a probabilistic
logic, where possible worlds are Herbrand interpretations for given
set of predicate letters $P$ in a given logic.  We use the mapping
\begin{equation} \label{eq>Montague}
I_n:\L_{op} \rightarrow \mathfrak{R}^{\W}
\end{equation}
 %in Definition \ref{def:MontagueIntens},
 where $\L_{op}$ is a subset of formulae with free variables (virtual predicates), such that for any virtual predicate $\phi(x_1,...,x_k) \in \L_{op}$ the mapping $I_n(\phi(x_1,...,x_k)):\W \rightarrow \mathfrak{R}$ is the Montague's meaning (i.e., \emph{intension}) of this virtual predicate \cite{Lewi86,Stal84,Mont70,Mont73,Mont74}, that is, the mapping which returns with the extension of this (virtual) predicate in every possible world in $\W$.
  The mapping $I_n$ can be
extended also to all sentences (the formulae without free
variables), such that for any sentence $\phi$, $I_n(\phi):\W
\rightarrow \{f,t\} = \P(\D^0) \subseteq \mathfrak{R}$ is a mapping
that defines the truth value (i.e., an extension in $\mathfrak{R}$) of this sentence in each possible world $\W$.
%Thus, for a given modal logic we will have that there is a bijection $$\F:\W \rightarrow \E$$ between the set of possible worlds and the set of  extensionalization functions.
 %
 \begin{definition} \cite{Majk11TS} \label{def:intensemant} \textsc{Two-step \textsc{I}ntensional \textsc{S}emantics:}
 The intensional semantics of the logic language with the set of formulae
$\L$ can be represented by the  mapping
\begin{center}
$~ \L ~\longrightarrow_I~ \D ~\Longrightarrow_{h \in \E}~\mathfrak{R}$,
\end{center}
where $~\longrightarrow_I~$ is a \emph{fixed intensional} interpretation $I:\L \rightarrow \D$ with image $im(I)\subset \D$, and $~\Longrightarrow_{h \in
\E}~$ is \emph{the set} of all extensionalization functions $h = \F(w):im(I) \rightarrow \mathfrak{R}$ in $\E$, where the mapping  $$\F:\W \rightarrow\E$$ is the bijection from the set of possible worlds (used by Montague)to the set of extensionalization functions.
\end{definition}
Based on this definition, we can establish the direct relationship between Bealer's and Montagues intensionality:\index{Bealer-Montague relationship}
%$\vspace*{-3mm}$
\begin{propo} \label{prop:Bealer-Montague} \textsc{Bealer-Montague relationship} \cite{Majk22}: \\
For any
logic formula (a virtual predicate) $\phi(\textbf{x})$, with a tuple of free variables $\textbf{x}$,  its extension in a possible
world $w \in \W$  satisfies the following equation
\begin{equation} \label{eq:Baeler-Montague}
~\F(w)(I(\phi(\textbf{x}))) = I_n(\phi(\textbf{x}))(w)
\end{equation}
\end{propo}

In a predicate logics, (virtual) predicates  expresses classes
(properties and relations), and sentences express propositions. Note
that classes (intensional entities) are \emph{reified}, i.e., they
belong to the same domain as individual objects (particulars). This
endows the intensional logics with a great deal of uniformity,
making it possible to manipulate classes and individual objects in
the same language. In particular, when viewed as an individual
object, a class can be a member of another class. In FOL we have the variables as arguments inside the predicates, and terms which can be assigned to variables are first-class objects while the predicates are the second-class objects. When we transform a virtual predicate into a term, by using intensional abstraction operator, we transform a logic formula into the first class object to be used inside another predicates as first-class objects. Thus, abstracted terms in the intensional FOL are just such abstracted terms as reification of logic formulae. For example, the sentence "Marco thinks \emph{that Zoran runs}", expressed by $thinks(\emph{Marco}, \lessdot runs(\emph{Zoran})\gtrdot)$ by using binary predicate $thinks$ and unary predicate $runs$ where the ground atom $runs(\emph{Zoran})$ is reified into the predicate $thinks$.

If $\phi(\textbf{x})$ is a formula (virtual predicate) with a list (a tuple) of free variables in $\textbf{x} =(x_1,...,x_n)$ (with ordering from-left-to-right of their appearance in $\phi$), and  $\alpha$ is its subset of \emph{distinct} variables,
 then $\lessdot \phi(\textbf{x}) \gtrdot_{\alpha}^{\beta}$ is a term, where $\beta$ is the remaining set of free variables  in $\textbf{x}$. The externally quantifiable variables are the \emph{free} variables not in $\alpha$. When $n =0,~ \lessdot \phi \gtrdot$ is a term which denotes a proposition, for $n \geq 1$ it denotes  a n-ary concept.

 We recall that we denote by $~t/g~$ (or $\phi/g$) the ground term (or
formula) without free variables, obtained by assignment $g$ from a
term $t$ (or a formula $\phi$), and by  $\phi[x/t]$ the formula \index{intensional abstraction} obtained by  uniformly replacing $x$ by a term $t$ in $\phi$.
\begin{definition} \label{def:abstrConv} \textsc{Intensional abstraction convention}:

 From the fact that we can use any permutation of the variables in a given virtual predicate,  we introduce the convention that
 \begin{equation}\label{eq:abstrctConv}
 \lessdot \phi(\textbf{x})\gtrdot_{\alpha}^{\beta}~~ is~a~ term~ obtained~ from~ virtual ~ predicate ~~\phi(\textbf{x})
 \end{equation}
 if $\alpha$ is \textsl{not empty}   such that  $\alpha\bigcup\beta$ is  the set of all variables in the list (tuple of variables)  $\textbf{x} = (x_1,...,x_n)$ of the virtual predicate (an open logic formula) $\phi$,  and $\alpha\bigcap\beta = \emptyset$, so that $|\alpha|+|\beta| = |\textbf{x}| = n$.
 Only the variables in $\beta$ (which are the only free variables of this term), can be quantified. If $\beta$ is empty then $\lessdot \phi(\textbf{x})\gtrdot_{\alpha}$ is a \emph{ground term}. If $\phi$ is a sentence and hence both $\alpha$ and $\beta$ are empty, we write simply $\lessdot \phi \gtrdot$ for this ground term.

 An assignment $g:\V \rightarrow \D$ forb variables in $\V$ is applied only to free variables in terms and formulae.  Such an assignment $g \in \D^{\V}$ can be recursively uniquely extended into the assignment $g^*:\T \rightarrow \D$, where $\T$ denotes the set of all terms (here $I$ is an intensional interpretation of this FOL, as explained in what follows), by :
\begin{enumerate}
  \item $g^*(t) = g(x) \in \D$ if the term $t$ is a variable $x \in\V$.
  \item $g^*(t) = I(c) \in \D$ if the term $t$ is a constant (nullary functional symbol) $c\in P$.
  \item If $t$ is an abstracted term obtained for an open formula $\phi_i$, $\lessdot \phi_i(\textbf{x}_i) \gtrdot_{\alpha_i}^{\beta_i}$,  then we must restrict the assignment to $g\in \D^{\beta_i}$ and to obtain recursive definition (when also $\phi_i(\textbf{x}_i)$ contains abstracted terms:
\begin{equation} \label{eq:assAbTerm}
  g^*(\lessdot \phi_i(\textbf{x}_i)\gtrdot_{\alpha_i}^{\beta_i}) =_{def}
    \left\{
    \begin{array}{ll}
   I(\phi_i(\textbf{x}_i))~~ \in D_{|\alpha_i|}, & \hbox{if  $\beta_i$ is  empty}\\
       I(\phi_i(\textbf{x}_i)/g)~~ \in D_{|\alpha_i|}, & \hbox{otherwise}
       \end{array}
  \right.
  \end{equation}
where $g(\beta) = g(\{y_1,..,y_m\}) = \{g(y_1),...,g(y_m)\}$ and $[\beta
/g(\beta)]$ is a uniform replacement of each i-th variable in the
set $\beta$ with the i-th constant in the set $g(\beta)$. Notice that $\alpha$ is the set of all free variables in the formula $\phi[\beta /g(\beta)]$.
\item  If $~t = \lessdot \phi_i\gtrdot$ is an abstracted term obtained from a sentence $\phi_i$ then \\$g^*(\lessdot \phi_i\gtrdot) = I(\phi_i) \in D_0$.\footnote{This case 4 replaces the same case in Definition 17 in \cite{Majk22} which can not happen because the terms are elements of the syntax of Intensional FOL and not the intensional concepts like $I(\phi(\textbf{x})) \in D_{|\textbf{x}|}$. This case 4 is the particular case 3 when the tuple of variables $\textbf{x}_i$ is empty and hence  $\beta_i$ and $\alpha_i$ are empty sets of variables with $|\alpha_i| = 0$.}
%\item $g^*(t) = I(\phi(\textbf{x})/g) \in D_0 \subset \D$ if  $t$ is an intensional term $I(\phi(\textbf{x})) \in D_{|\textbf{x}|}$. \footnote{It happen when the abstracted term (\ref{eq:assAbTerm}) is inside another predicate and we make assignment for attributes of this predicate.}
\end{enumerate}
The abstracted terms $\lessdot \phi \gtrdot_\alpha^\beta$ can be used as terms in any predicate $p_j^k\in P$, for example for an atom $p^3_j(\lessdot\phi \gtrdot_\alpha^\beta,y,z)$ with free variables $y$, $z$ and that in $\beta$.
Let $p^k_j(t_1,...,t_k)$ be an atom  with at least one of abstract term $t_i = \lessdot \phi_i(\textbf{x}_i)\gtrdot_{\alpha_i}^{\beta_i}$ with $\beta_i$ non empty and let $\beta$ denotes the union of all $\beta_i$ of the abstracted terms in this atom. We can consider this atom  as a virtual predicate $\phi(\textbf{x})$ with ordered tuple of free variables $\textbf{x}$, and  we denote by $\textbf{y}$ its ordered subtuple  without variables in $\beta$ with $1\leq n= |\textbf{y}| $.  Then we have that for each assignment $g \in \D^\beta$, $p^k_j(t_1/g,...,t_k/g)$ is a standard atom (all abstracted terms $t_i/g = g^*(t_i)$ by using (\ref{eq:assAbTerm}) are transformed to values in $\D$ and $I(p^k_j(t_1/g,...,t_k/g))\in D_{n} \subset \D$), while for Tarskian interpretation we obtain the following set of tuples:
\begin{equation} \label{eq:IntAbstTerm2}
I_T^*(p^k_j(t_1/g,...,t_k/g)) =_{def} \{g_1(\textbf{y})\mid g_1\in \D^{\overline{\textbf{y}}} ~~\emph{and}~~ I_T^*(p^k_j(t_1/g,...,t_k/g)[\overline{\textbf{y}}/g_1(\overline{\textbf{y}})])= t\}
\end{equation}
where $I_T^*(p^k_j(t_1/g,...,t_k/g)[\overline{\textbf{y}}/g_1(\overline{\textbf{y}})])= t~~$ iff $~~(t_1/g,...,t_k/g)[\overline{\textbf{y}}/g_1(\overline{\textbf{y}})] \in I_T(p^k_j)$, and we recall that $I_T^*(p^k_j(z_1,...,z_k)) =_{def}I_T(p^k_j)$ if all $z_i$, $1\leq i\leq k$, are free variables.\\
 So, general  Tarski's and intensional interpretations, when we have at least one hidden variable in the terms, are defined by
\begin{multline} \label{eq:IntAbstTerm}
I_T^*(p^k_j(t_1,...,t_k)) =_{def} \bigcup_{ g \in \D^\beta }I_T^*(p^k_j(t_1/g,...,t_k/g))~~\neq~I_T(p^k_j) \subseteq \D^k\\
I(p^k_j(t_1,...,t_n)) =_{def} union(\{I(p^k_j(t_1/g,...,t_k/g)) \mid g \in \D^\beta \})~~~~~~~~~~~~~~~~~~
\end{multline}
where derived operator "union" of intensional algebra is provided in \cite{Majk22}.
\end{definition}
\textbf{Remark}: We recall that the assignment for abstracted terms is defined in this way in order to eliminate all abstract operators with such assignments, as required by Proposition 7 in \cite{Majk22}.  With such definition we forget the natural-language structure generated by reification (by using abstracted operators). We can change this definition of assignment for abstracted terms if we do not want to forget the natural-language expression structure during the assignments.
The \emph{choice} used in Definition \ref{def:abstrConv} of the Intensional abstraction convention, adopted in \cite{Majk22} has been motivated by the interest to obtain a conservative evolution from the standard RDM database systems in more expressive Intensional RDB (IRDB) with open schemas, many-valued attributes of relational tables and the possibility to manage both the data and the metadata in the same SQL-like database languages. So, it is avoided to have the abstract terms in the PRP domain $\D$ of the particulars and the concepts, and in the standard $\D$-domain based collection of relations
  $~\mathfrak{R} = \bigcup_{k \in \mathbb{N}} \P(\D^k) = \sum_{k\in \mathbb{N}}\P(D^k)$.\\
Such limitation can be avoided, in particular for the AGI (General Artificial Intelligence) applications, strongly oriented to the natural language manipulations.\\
Moreover, we need to render explicit definition of the intensional interpretation of the ground atoms (with all terms without variables and also the abstracted terms, which, differently from the definition in (\ref{eq:IntAbstTerm}), have no  any hidden variable, by
\begin{equation} \label{eq:IntAbstTerm2noHidden}
I(p^k_j(t_1,...,t_k)) =_{def} I(p^k_j(t_1,...,t_k))/g = I(p^k_j(g^*(t_1),...,g^*(t_k)))\in \D_0
\end{equation}
for any given assignment $g\in \D^\V$.
\\$\square$\\
To show how the abstracted terms are very powerful natural language properties, which can be used in AGI (for the strong-AI robots \cite{Majk23r}), let us consider the following example:
\begin{example} \label{ex:2}
Let us consider the 'Know' predicate $p_1^3(x_1,x_2,t_1)$ used for epistemic reasoning for the robots in \cite{Majk24a}), where the variable $x_1$ is used for the  grammatical-form of the verb   "to know" (time, gerund, etc..)$x_1$ (with values, for example, "past", "present", "future" for time-specification, which are the IFOL constants $c_i$ and particulars in $D_{-1}$, so that for any assignment $g$, $g(c_i) = c_i \in \D$), the variable $x_2$ is used for the subject of this knowledge and $t_1$ is used for an abstracted term expression for this particular knowledge), and the built-in predicate $p^7_j$ "sphere-formula"  (corresponding to the standard preorder binary relationships $\leq$ for reals) with the atom containing  the free variables $p^7_j(x,x_0,y,y_0,z,z_0,v)$ representing the algebraic equation $(x-x_0)^2+(y-y_0)^2+(z-z_0)^2 \leq v^2$ where $x_0,y_0,z_0$ are coordinates of the center of a sphere with radius $v$ (used in Examples 7 and 8 in \cite{Majk22}, Chapter 2 for Constrained databases), and the
following natural language expression:\\

"Somebody  knows that his friend told  that $(x-x_0)^2+(y-y_0)^2+(z-z_0)^2 \leq v^2$ is a sphere-formula with the center in $(x_0,y_o,z_0)$ and radius $v$"\\\\
which is parsed into the following IFOL \emph{open} formula (in the example in next section we will preset this process of parsing) because the word "somebody" referred to some person is an variable in a predicate-based logic:
\begin{equation} \label{eq:NLsphere}
p_1^3(present, x_2, \lessdot p_2^3(past, x_4, \lessdot \preccurlyeq^2((x-x_0)^2+(y-y_0)^2+(z-z_0)^2,v^2)\gtrdot^\beta_\alpha) \gtrdot^{\beta\bigcup\{x_4\}}_\alpha)
\end{equation}
where $\preccurlyeq^2$ is the binary built-in predicate such that for any assignment $g$, the ground atom  $\preccurlyeq^2((x-x_0)^2+(y-y_0)^2+(z-z_0)^2,v^2)/g$  is true iff $(g(x)-g(x_0))^2+(g(y)-g(y_0))^2+((g(z)-g(x_0))^2 \leqslant g(v)^2$. The $p_2^3(x_3,x_4, t_2)$ is the atom of the predicate "to tell" with the variable $x_4$ for who is subject that tells, $x_3$ (as $x_1$)   is used for the  grammatical-form of the verb   "to tell" (time, gerund, etc..) with values, for example, "past", "present", "future" for time-specification,
and abstracted term $t_2$ for what is told, and where $\beta = \{x_0,y_0,z_0,v\}$ are the visible (free) variables and $\alpha = \{x,y,z\}$ are hidden variables of the constrained (compressed) database representation of a sphere.  The variable $x_2$ represents "somebody" while $x_4$ represents "a friend of somebody".\\
Thus, in the logic formula (\ref{eq:NLsphere}) we have the abstracted terms and the set of free variables $\V = \{x_2,x_4\}\bigcup \beta$, so we can consider the following assignment $g \in \D^\V$, with \\
- $~g(x_2) = "Zoran ~Majkic"$\\
- $~g(x_4) = "Alberto ~Rossi"$, a friend of Zoran Majkic\\
- for variables of real in $\beta$, $g(x_0) = g(y_0) = g(z_0) = 0.0$ and $g(v) = 2.0$, for a sphere with center at the coordinate origin and with radius 2.0. Note that the tuple of variables in $\alpha$, $(x,y,z)$, represents a point inside this sphere.

Consequently, this assignment $g \in \D^\V$ extends to all terms in the IFOL open formula (\ref{eq:NLsphere}) as follows (by $\phi(\textbf{x})$ we denote the atom $\preccurlyeq^2((x-x_0)^2+(y-y_0)^2+(z-z_0)^2,v^2)$):\\
$p_1^3(present, x_2, \lessdot p_2^3(past, x_4, \lessdot \phi(\textbf{x})\gtrdot^\beta_\alpha) \gtrdot^{\beta\bigcup\{x_4\}}_\alpha)/g =\\
= p_1^3(present, g^*(x_2), g^*(\lessdot p_2^3(past, x_4, \lessdot \phi(\textbf{x})\gtrdot^\beta_\alpha) \gtrdot^{\beta\bigcup\{x_4\}}_\alpha))\\
= p_1^3(present, g(x_2), I( p_2^3(past, x_4, \lessdot \phi(\textbf{x})\gtrdot^\beta_\alpha)/g))\\
= p_1^3(present, g(x_2), I( p_2^3(past, g(x_4), g^*(\lessdot \phi(\textbf{x})\gtrdot^\beta_\alpha))))\\
= p_1^3(present, g(x_2), I( p_2^3(past, g(x_4),\\
 %\lessdot\preccurlyeq^2((x-g(x_0))^2+(y-g(y_0))^2+(z-g(z_0))^2,g(v)^2)\gtrdot_\alpha)))\\
I(\preccurlyeq^2((x-g(x_0))^2+(y-g(y_0))^2+(z-g(z_0))^2,g(v)^2))))\\
%= p_1^3(in~present, g(x_2), I( p_2^3(in~past, g(x_4), \lessdot \preccurlyeq^2(x^2+y^2+z^2,4.0)\gtrdot_\alpha)))$\\\\
= p_1^3(present, g(x_2), I( p_2^3(past, g(x_4), I( \preccurlyeq^2(x^2+y^2+z^2,4.0))))$\\\\
which is a ground atom (a sentence) with last argument the intensional 0-ary concept (a proposition), $u = I( p_2^3(past, Alberto ~Rossi, I( \preccurlyeq^2(x^2+y^2+z^2,4.0)))) \in D_0$
%with the ground abstract term (without free variables) $\lessdot \preccurlyeq^2(x^2+y^2+z^2,4.0)\gtrdot_\alpha$,
such that  $u_2= I(I(\preccurlyeq^2(x^2+y^2+z^2,4.0))) \in D_3$ is a 3-ary intensional concept whose extension is \\
$h(u_2) = \{(g'(x),g'(y).g'(z))\mid ~for ~any ~g' \in \D^\alpha \\~such ~that~ \preccurlyeq^2(x^2+y^2+z^2,4.0)/g'~ is~ true~\}\\
= \{(g'(x),g'(y).g'(z))\mid ~for ~any ~g' \in \D^\alpha ~with~
g'(x)^2+g'(y)^2+g'(z)^2 \leq 4,0\} $\\
which is just the collection of all points of this sphere, where
the values $g'(x)$, $g'(y)$, $g'(z)$, $4.0$ are the particulars in $D_{-1}$, which are elements of the unary built-in concept $Reals \in D_1$, that is, for example, has to be satisfied that $ g'(x) \in h(Reals)$ for each extensionalization  function $h$. So, in many-sorted version of IFOL the concept $Reals \in D_1$ will be a sort of this variable $x$, and this constraint would be automatically applied.

Notice that the intensional concept $u$ above is the propositional concept corresponding to \\\\
"Alberto Rossi told that $x^2+y^2*z^2 \leq 4.0$ is a sphere-formula with the center $(0.0,0.0,0.0)$ and radius $2.0$"\\\\
while the intensional concept \\$u_3 =I(p_1^3(present, g(x_2), I( p_2^3(past, g(x_4), I(\preccurlyeq^2(x^2+y^2+z^2,4.0)))))) \in D_0$ is the propositional concept corresponding to \\\\
"Zoran Majkic knows that his friend Alberto Rossi told that $x^2+y^2*z^2 \leq 4.0$ is a sphere-formula with the center $(0.0,0.0,0.0)$ and radius $2.0$"\\\\
which is just a particular subcase of the initial more general natural language expression. So, with this example we have shown how the Intensional FOL is much more powerful, w.r.t the standard FOL, to express the natural language meaning by using the intensional concepts and preserving the Tarski's truth-semantics for the logic sentences.\\
$\square$
\end{example}
Let us consider another example of a natural language \emph{sentence} with abstracted terms obtained also from non Bealer's "that-syntax" :
\begin{example} \label{ex:3}
Le us consider the following natural language expression where "EN" means "Euclid Number":\\

"Mario Rossi works to resolve the EN-problem for which the people do not believe there exists somebody who resolved it."\\

which, after the parsing into IFOL predicate-based syntax, uses the following three predicates presented by their logic atom syntax with  variables and abstracted terms:\\
- $~p_1^3(x_1, x_2, t_1)$  for the verb "to work", where the time-variable $x_1$ is used for the  grammatical-form of the verb   "to work" (time, gerund, etc..) with values, for example, "past", "present", "future" for time-specification,  the variable $x_2$ is used for the subject who is working and $t_1$ is used for an abstracted term expression for this particular job.\\
- $~p_2^3(x_1, x_2, t_2)$  for the verb "to resolve", where the time-variable $x_1$ is used for the  grammatical-form of the verb   "to resolve" (time, gerund, etc..) with values, for example, "past", "present", "future" for time-specification,, the variable $x_2$ is used for the subject who is resolving and $t_2$ is used for an abstracted term expression for this particular problem to be resolved.\\
- $~p_3^3(x_3, x_4, t_3)$  for the verb "to believe", where the time-variable $x_3$ is used for the  grammatical-form of the verb   "to believe" (time, gerund, etc..) with values, for example, "past", "present", "future" for time-specification,, the variable $x_2$ is used for the subject who believes and $t_3$ is used for an abstracted term expression for this particular believing.\\

Thus, the logic sentence of IFOL obtained by such parsing has to be representing by the following logic formula, where "Mario Rossi", "EN-problem" and "people" are logic constants $c_i$ (nullary functions in IFOL) which are in $\D$ as well, so that for any assignment $g$, $g(c_i) = c_i$:
\begin{multline} \label{eq:NLsentence}
p_1^3(present, Mario~ Rossi, \lessdot p_2^3(present, Mario~ Rossi, EN-problem)\gtrdot)\\
\wedge \neg p_3^3(present, people, \lessdot (\exists x_2)p_2^3(past, x_2, EN-problem)\gtrdot)
\end{multline}
The logic conjunction $\wedge$ of two ground atoms of the predicate $p_1^3$ in the first line, and $p_3^3$ in the line bellow, corresponding to natural language sentences

(a) "Mario Rossi works to resolve the EN-problem", and

(b) "The people do not believe that there exists somebody who resolved it (refered to the EN-problem)",\\
\\
where the expression  "there exists somebody" in the  second line sentence is logically represented by the existentially quantified variable $\exists x_2$.
 \\$\square$
\end{example}
This paper is also a continuation of  my approach to AGI (Strong-AI) for a new generation of intelligent robots using \emph{natural languages with epistemic logic reasoning}, recently published in the papers \cite{Majk23r} and \cite{Majk24a}. Neuro-symbolic AI attempts to integrate neural and symbolic architectures in a manner that addresses strengths and weaknesses of each, in a complementary fashion, in order to support robust strong AI capable of reasoning, learning, and cognitive modeling. In this approach to AGI I considered the Intensional First Order Logic (IFOL) \cite{Majk22} as a symbolic architecture of modern robots, able to use natural languages to communicate with humans and to reason about their own knowledge with self-reference and abstraction language property.

 This paper is dedicated to the extension of this previous work and to the more natural relationship between the PRP concepts, as the fundamental parts of the natural language vocabulary, and the First-order Logic based on abstract mathematical concept of the n-ary predicates introduced as mean how to compute the truth of logic formulae. Differently from natural ideas of the concepts as a human mental objects used in natural languages, the mathematical structure of predicates, useful for the computation of the truth of logic formulae and logic reasoning (deduction), are abstract mathematical structures that human does not use in every-day communication and reasoning.  The most general and natural enrichment of such abstract mathematical logic structures of formal extensional logics is to use these intensional concepts, which are dominant part of Intensional FOL as types (sorts) to assign to the variables used to define the terms and the logic atoms of the predicates.  Thus, differently from the previous development of \emph{unsorted} Intensional FOL summarized in my  recent book \cite{Majk22}, in this paper we will develop the theory of \emph{Many-Sorted} Intensional FOL, more sophisticated and human-mentally  oriented  language approach. The formal mathematical complication of this many-sorted version of the IFOL will be balanced by a more effective and controlled semantics, by limiting the assignment of the values to each variable of the formal logic only to the strict subset of the values of the domain that is just the extension of the intensional concepts used as the sorts of this variable.

\section{Many-Sorted Intensional First-Order Logic}
Research on ontology has been increasingly widespread in the computer science community from 1990 and its importance is being recognized in diverse research fields, in my case in knowledge representation \cite{Gru95,GuMV98}, database design and database integration, information modeling, information retrieval and object-oriented analysis (European SEWASIE Project, University La Sapienza, Rome), in the period 1998-2008.

 In particular, the IFOL (Intensional First-order logic \cite{Majk22} supports by its PRP intensional concepts the intended \emph{meaning} (sense) and grounding of them to the reality by experience and learning, so it is adequate to support the development of AGI robots. Tom Gruber defined an ontology as "a specification of a conceptualization" \cite{Gru95} as a structure $(\D,\textbf{R})$: then Guarino  discussed such a definition in \cite{Guar98}, arguing that, in order for it to
have some sense, a different, \emph{intensional} account of the notion of conceptualization has to
be introduced, by considering Montague's definition of intension with the set of possible worlds $\W$ and hence the triple $(\D,\W,\R)$ of conceptualization where $\R$ is the set of  conceptualized relations (the functions from $\W$ to their admittable extensions). In this way, the structure of Gruber above is in reference to a particular possible world $w \in \W$.

 So, w.r.t. Guarino's conceptualization (his "ontological commitment" corresponds to our intensional interpretation $I$ which assigns the particulars of  $D_{-1}$ to the constant symbols (nullary functional symbols in $P$ of FOL) and intensional concepts (universals) of $D_1+...+ D_n \subset \D$ to the atoms of predicates with free variables for all its arguments),  in our case $\D$ is the PRP domain (of particulars $D_{-1}$ and and universals $D_I$) and $\R$ is a set of n-ary ($n\geq 1$) intensional concepts in $D_1+...+ D_n \subset \D$. Thus, we can use Guarino's definition of the relationships between vocabulary, conceptualization, ontological commitment and ontology, expressed here by the Bealer-Montague intensionality relationship (\ref{eq:Baeler-Montague})
 $$~\F(w)(I(\phi(\textbf{x}))) = I_n(\phi(\textbf{x}))(w)$$
 for each virtual predicate $\phi(\textbf{x})\in \L_{op}$ (an open formula in $\L_{op} \subset \L$ of FOL syntax), Montague's possible world $w \in \W$ and mapping $I_n$ in (\ref{eq>Montague}) such that $I_n(\phi(\textbf{x})):\W\rightarrow \mathfrak{R}$ is Montague's meaning (intension) of this virtual predicate, and fixed "ontological commitment" (intensional interpretation $I$).
%---------------

We are interested especially for two kinds of relationships between intensional concepts in $D_I$: the intensional equality of  concepts $'='$ (in the intensional algebra of  concepts $\A_{int}$, provided by (1.20) of Example 4 in \cite{Majk22}) with Montague's intension definition based on possible worlds (in bijection with the set of extensionalization functions $h= \F(w)\in \E$) in Proposition \ref{prop:Bealer-Montague} of previous Section,  and the IS-A relationship $'\sqsubseteq'$ which generalized IFOL semantics (\emph{non restricted to unary concepts} only as in DL, but between any two intensional concepts in $D_1+D_2+...+D_n$ for enough big finite natural number $n\geq 1$, independently of their arity) will be defined in what follows.

In IFOL the concepts can have, attached to them, a number of sorts, so that each concept $u \in D_k$ (a k-ary Relation concept in PRP), with $k\geq 1$, has \emph{a name} (expressed by natural language lexicon in bpld letters) and exactly $k$ sorts $s_i$ attached to it (which are also the concepts of the PRP domain $\D$).
The C-structures,  that we introduce by IFOL, are the part of ontology of the concepts  in $D_n \subset \D, n\geq 1$ (Properties an Relations in PRP domain $\D$), where one of the basic relationship is the IS-A relationship between the concepts so that the algebraic expression of concepts, $u_1 \sqsubseteq u_2$ means that the concept with the name $u_1$ is a part of the more general concept with name $u_2$.

 We can avoid the \emph{logical }definition of the ontologies by using the axioms of the First-order logic, but to build the ontology directly on the PRP domain of intensional concepts using at least these two relationship in enriched intensional algebra $\A_{int}$ by these two relationships expressed by the symbols $'='$ and $'\sqsubseteq'$.

%-------------------------------------------------------------
It is a common assumption that predicates stand for properties. However, there are different views of how to understand the semantic relation of 'standing for' and how to understand the properties predicates are supposed to stand for ontologically. Frege took predicates to stand for concepts, unsaturated entities, rather than objects, the denotations of all referential terms (including explicit concept-referring and property-referring terms).  There are also philosophers such as George Bealer \cite{Beal82} who take predicates to stand for properties as saturated objects, with a predication relation ensuring the relation of a property
to its bearer.

However, here we make a clear distinction between the human-mind concepts expressed by the \emph{natural language} lexicon phrases (composed by more than one words as well), so that in PRP domain $\D$ of intensional FOL, each n-ary concept (Property as unary concept and Relations as n-ary concepts for $n\geq 2$) is just a particular natural language single word (in the simplest case), or composed phrase by a number of words, while the \emph{predicates}, identified by special logic symbols $p^k_j$ are the \emph{mathematical abstraction} which are not used in normal human communication but in the mathematical logic useful for formal computation of the truth (in our case, by generalized Tarski's interpretation $I_T^*:\L\rightarrow \mathfrak{R}$ of the \emph{formal logic formulae} of the Intensional FOL).

 Just because of such an explicit difference, in this Section we need to specify a kind of the concepts that can be transformed in formal IFOL by the atoms  of a corresponding predicate $p^k_j \in P$ to be used for logic epistemic reasoning. Such subset of the PRP concepts that are good candidates to be used in IFOL as the formal predicates, will be denominated as "predicate-concepts".
 \\
 \textbf{Remark}:  It is well known that the predicates of the standard FOL are extensional entities because the standard Tarski semantics define a relation (extension) to each predicate in FOL.  So it seems contradiction that we can assign to the \emph{intensional} predicate-concepts the extensional entities. But it only seems to be a contradiction, because in Intensional FOL, also the predicate symbols are intensional entities, and we can define the "Tarskian possible worlds" by (see (1.9) in \cite{Majk22}) corresponding to the Montague's posible worlds $w\in \W$, for the given "ontological commitment" (intensional interpretation) $I$,  by
 $$\W_e =_{def} \{I_T^* = h\circ I = \F(w)\circ I \mid w\in \W\}$$
 and hence, while the intensional concepts in IFOL have their intension (meaning) defined w.r.t the set of the possible worlds $h\in \E$,  the predicate symbols of IFOL are intensional entities w.r.t. the set of possible worlds in $\W_e$. The predicate symbol in IFOL becomes an "extensional" entity only in some specific world $I_T^* \in \W_e$, that is in some specific Tarski's interpretation. More precisely, each atom of a given predicate (with free variables) is an intensional entity in IFOL, which in given specific world $I_T^* \in \W_e$ has a precise extension (relation) determined by this world.\\
 $\square$\\
  In this way, by assigning a value to any free variable of the predicate $p^k_j$ its corresponding concept will be just a subconcept of this predicate-concept, with implicit transitive relationship IS-A between them.   In this way, for a given predicate $p^k_j$ of IFOL, we will have a family of the concepts in PRP (expressed as composed natural language words), such that each concept of this family corresponds to a particular assignment $g$ to the variables of the logic atom of this predicate.
  \begin{definition} \label{def:var-sort}
  We assume the UNA (Unique Name Assumption) for all concepts in the PRP domain $\D$, and hence that a \emph{sort} is identified by the name (in bold letters, defined as a natural language phrase)  of corresponding concept.\\
     The syntax of a k-ary  concept  $u \in D_k$, $k\geq 1$,  is composed by its name  followed by $k\geq 1$  list of its sorts $s_i$, which are the concepts in $D_1+D_2+...+D_n $ or the special sort $'nested ~
 sentence'$ (which is not an element of the  domain $\D$), representing its properties, i.e., a concept $u\in D_k$ has a syntax form $\textbf{phrase}:s_1,...,s_k$.\\
 %where the \textbf{phrase} is called "the name of the concept" and the sorts $u_i \in D_1+D_2+...+D_n$ are its attributes.
 So, if this concept $u$ is a predicate-concept (which defines a  k-ary predicate letter $p_i^k \in P$ of the FOL), there is the following relationship between the atoms of the predicates with terms $t_i$ (in this definition of predicate-concept we do not use for $t_i$  a constant (a nullary function) but a variable, in order to obtain a more general concept corresponding to this predicate letter $p_i^k \in P$),
  and corresponding predicate-concepts with the list of \emph{sorts}  $s_i$ in $\S \subset D_1+D_2+...+D_n+ \{nested ~
 sentence\}$,
 \begin{equation} \label{eq:P-Ccorrespondence}
  p_i^k(t_1,...,t_k) ~~~~~\Leftrightarrow~~~~~ \textbf{phrase}:s_1,...,s_k \in D_k
  \end{equation}
  such that if a term $t_i$ is a free variable then $s_i$ is the sort of this variable,  if $t_i$ is a functional term then $s_i$ is the sort of term's outermost functional symbol, if $t_i$ is an abstracted term then $s_i$ is the $'nested ~
 sentence'$ sort\footnote{Such abstracted term are inserted in the predicate with this variable during the parsing of a complex natural language sentences containing the nested subsentences. So, the sort $'nested ~ sentence'$, which is not an element of a domain $\D$, has the empty extension in domain $\D$ differently from all other sorts.}.\\
  Thus, we obtain the following mapping for the sorts of variables:
 \begin{equation} \label{eq:var-sort}
 \digamma:\V \rightarrow \S
 \end{equation}
   Let us define how this mapping apply to the predicate-based IFOL syntax:\\
 1. For each variable $x \in \V$, to which we assign a sort $s \in S$ (different from 'nested sentence'), denoted by $x:s$, we have that  $\digamma(x) = s$ with non-empty extension $\|s\|$.\\
 2. If the term $t_i$ is an abstracted term, then $\digamma(t_i) = 'nested ~sentence' \notin \D$.\\
 3.  For the logic constant $c$ (a nullary functional symbol) the $\digamma(t_i)$ is its defined sort.
 Otherwise,  term’s sort is the sort declared as return sort of the term’s outermost function symbol, that is, for k-ary functional symbol $f^k_i$ with $k\geq 1$ the term  $t = f_i^k(t_1,...,t_k)$ which represents function\\
 $f_i:\|\digamma(t_i)\|\times...\times \|\digamma(t_i)\| \rightarrow \|s\|$\\
 where $s \in S$ is the sort of this functional term, that is, $\digamma(t) = s$. From the fact that in IFOL the functional symbols are represented as particular predicates, we obtain the sort of $f^k_i$ as

  $\digamma(f^k_i) = \digamma(t_1)\times...\times\digamma(t_k)\times s$.\\
 4. For each predicate symbol $p^k_i \in P\backslash F$, $k\geq 1$, corresponding to the predicate-concept $phrase:s_1,...,s_k$, its sort is

 $\digamma(p^k_i) = s_1\times...\times s_k$\\
5. For each virtual predicate $\phi_i$ (an open FOL formulae with the non empty tuple of free variables $\textbf{x}= (x_1,...x_k)$, $k\geq 1$), corresponding to composed concept by the operators in $\A_{int}$ in \cite{Majk22}, its sort is

 $\digamma(\phi_i) = \digamma(x_1)\times...\times\digamma(x_k)$.
 \end{definition}
 For any two sorts (concepts), $s_1,s_2 \in S$, with IS-A relationship $s_1 \sqsubseteq s_2$ we say that $s_2$ is a \emph{supersort} (superconcept) of $s_1$, and $s_1$ is a \emph{subsort} (subconcept) of $s_2$. To make it clear that
we mean this notion of sort, we will talk about the \emph{static sort} of a term.
\begin{definition} \label{def:Static Sorts} \textsc{Static Sorts}:
Every element of the domain $D$ has exactly one sort, defined as follows, following the PRP domain $\D$ of the IFOL:
\begin{enumerate}
  \item For any particular $u\in D_{-1}$, its sort is $\digamma(u) \in D_1$;
  \item For any proposition $u\in D_0$, its sort is $\digamma(u)=\textbf{truth values}:\textbf{truth values} \in D_1$, which for a standard FOL has extension of two truth values, $\|\textbf{truth values}\| =h(u)=\{f,t\}$;
  \item For a property (unary concepts), $u = \textbf{phrase}:s \in D_1$, its sort is  $\digamma(u) =s$, with the extension $\|\digamma(u)\| = \|s\|=h(u)$;
  \item For a k-ary, $k\geq 2$, relational concept $u = \textbf{phrase}:s_1,...,s_k \in D_k$, used as a (single) sort $s$, such that $\digamma(u) = \digamma(\textbf{phrase}:s_1,...,s_k) = s$, which extension $\|s\|$ is derived from the IS-A tree $T_u$ of the subconcepts of $u$ as a set containing the name of each relational concept (intermediate nodes of this ontology tree) and the extensions  of the leafs of this tree (which are unary concepts), that is,\\
      $\|\digamma(u)\| = \|s \| =
      \{\textbf{phrase}_i\mid \textbf{phrase}_i$ is the name of a m-ary concept with $m\geq 2$, such that $\textbf{phrase}_i \sqsubseteq \textbf{phrase}\} ~
      \bigcup_{\textbf{phrase}_j:s_j \in L}\{\textbf{phrase}_j\circ u_n \mid u_n \in \|s_j\|\}$,\\
      where $L$ is the set of  unary subconcepts  which are the leafs in the tree $T_u$, and  $\textbf{phrase}_j\circ u_n$ denotes a string (phrase) obtained by concatenation of the strings $\textbf{phrase}_j$ and $u_n$.  Notice that the extension of the sort of the concept  $u$ is a set of strings, while its extension as an intensional k-ary relation is the set of k-ary tuples in $h(u)$.
\end{enumerate}
From the fact that we do not use the sorts of all concepts in $\D$ we define the finite subset of sorts derived by the concepts in $\D$ by $S_D$, and used by IFOL in specific applications, and the set of all sorts by $S =S_D \bigcup \{nested~ sentence\}$ where '\emph{nested sentence}' sort is not an element of the PRP domain $\D$, but used as a sort of all abstracted terms used in predicate atoms. The extension of this sort are the nested sentences of a natural language, used inside other natural language sentences.
\end{definition}
 So, the extension of any sort  corresponding to a concept $u \in D_k$, $k\geq 1$, is standard set of values (an unary relation)\footnote{For example, consider unitary basic concepts \textbf{reals}:\textbf{reals} of the sort \textbf{reals}, \textbf{integers}:\textbf{integers} of the sort \textbf{integers}, \textbf{date}:\textbf{date}  of the sort \textbf{date} with fixed extensions $\|\textbf{reals}\| = h(\textbf{reals}:\textbf{reals}) = \mathbb{R}$ (the set of real numbers), $\|\textbf{integers}\| =h(\textbf{integers}:\textbf{integers}) = \mathbb{Z}$ (the set of integer numbers),  and their subconcepts $\textbf{closed interval [0,1]}:\textbf{reals}$ with the extension  $\|\textbf{closed interval [0,1]}\| =h(\textbf{closed interval [0,1]}:\textbf{reals}) = [0,1]\subset \mathbb{R}$,  and for a subconcept \textbf{weekdays}:\textbf{date} with $\|\textbf{weekdays}\| =h(\textbf{weekdays}:\textbf{date}) \subset \|\textbf{date}\| =h(\textbf{date}:\textbf{date})$.
 }.
  Because of that, for the concepts used in IS-A relationships $'\sqsubseteq'$ we will use only the names of the concepts without their list of sorts. That is, any two concepts in IS-A relationship $u_1\sqsubseteq u_2$, independently of their arity, are considered as the sorts (with unary relations as their extensions) so that we obtain the \emph{inclusion} relationship between them of  their unary relations.\\
\textbf{ Remark:} Note that the predicate-concept correspondence is partially represented by the intensional interpretation $I$, that is (\ref{eq:P-Ccorrespondence}) satisfies the mapping

  $\textbf{phrase}:s_1,...,s_k ~=~I(p_i^k(t_1,...,t_k))$,\\ and analogously for all subconcepts of this predicate-concept.\\
  However, in the intensional algebra of concepts we can have also other relationships between concepts that are not intensional interpretations of the FOL connectives. In fact for the IS-A relationship $'\sqsubseteq'$ between the concepts we do not provide any binary predicate in IFOL, no relative axiom. We also want that the intensional structures remain open for another kinds of the knowledge relationships between the concepts, in order to obtain more expressive C-structures (schemas): A schema (plural form: schemata) is a structured cluster of concepts, it can be used to represent objects, scenarios or sequences of events or relations. The philosopher Immanuel Kant first proposed the concept of schemata as innate structures used to help us perceive the world.

   This is based on the human mind in which the logic-deductive component of human reasoning is only a part of the whole conceptual knowledge (schemata) learned during the life by personal experience.
   \\$\square$\\
 Let us show this mechanism by a simple example, by using a following part of the IS-A relationship of the concept's ontology lattice:  $\textbf{cat} \sqsubseteq \textbf{animal}$, $\textbf{dog} \sqsubseteq \textbf{animal}$, $\textbf{wolve} \sqsubseteq \textbf{animal}$, $\textbf{fox} \sqsubseteq \textbf{animal}$, $\textbf{bird} \sqsubseteq \textbf{animal}$, $\textbf{caterpillars} \sqsubseteq \textbf{animal}$, $\textbf{snail} \sqsubseteq \textbf{animal}$, and consider in details first one.  It can be expressed in a more detailed ontology structure of the concepts  $\textbf{cat} \sqsubseteq \textbf{animal with brests} \sqsubseteq \textbf{animal}$.\\
 From the fact that the concepts $\textbf{cat}:s'_1,...,s'_k$ and $\textbf{animal}:s_1,...,s_n$ are expressed by a single word name (thus non specified by other properties), they are proper candidates to be logically expressed by two different predicates and their atoms with free variables $p_j^k(x_1,...,x_k)$ and the atom $p_m^n(y_1,...,x_n)$ relatively, and suppose that the variable $y_1$ specifies if the animal is with breasts or not, that is by specifying that the sort of the variable $y_1\in \V$ is the concept $\textbf{kind of animals}$ , i.e, from (\ref{eq:var-sort}), $s_1 =\digamma(y_1) = \textbf{kind of animals} \in \S$ so that the possible value in the extension of this sort is  \textbf{with breast}
  %$\in h(\textbf{kind of animals})$
  as well, i.e,. $\textbf{with breast} \in \|s_1\|$, and we denote by $g$ the assignment such that assigns a value only to the free variable $y_1$,  $g(y_1) =  \textbf{with breast}$.  Thus, in this case the subconcept (w.r.t. the predicate-concept $\textbf{animal}:s_1,...,s_n$ in $ D_{n}$ ),  $\textbf{animal with breasts}:s_2,...,s_n$ in $D_{n-1}$, logically is expressed by the atom $p_m^n(g(y_1),y_2,,...,y_n)$ in which we have no more the free variable $y_1$ because its value is fixed.\\
  Equivalent classes of the subconcepts:
  Let $s_2$ be the sort \textbf{hairness} with extension $\|s_2\| =\{hairy, hairless\}$, and the assignment $g$ assigns also to the variable $y_2$ of the predicate $p_m^n$ the value $g(y_2) =  hairy$, then we obtain two subconcepts

   $u_3 = \textbf{animal with breasts hairy}:s_3,...,s_n$ in $D_{n-3}$ and

    $u_4 = \textbf{animal hairy with breasts}:s_3,...,s_n$ in $D_{n-3}$
    \\which is the same as the subconcept $u_3$, from the fact that both are corresponding to the \emph{same} predicate atom $p_m^n(g(y_1),g(y_2),,...,y_n)$, with only permuted attributes in the names of these two subconcepts. From the fact that in this way for the same predicate-concept, we can have a number of equivalent subconcepts with only different permutation of the attributes in their subconcept names.
 In order to preserve the bijection between predicate' atoms and the subconcepts of given predicate-concept, we will use only the subconcept's name in which the attributes are ordered as corresponding variables of the predicate: we will call such subconcepts as \emph{canonical subconcepts}. So, in our example, we will not generate the subconcept $\textbf{animal hairy with breasts}:s_3,...,s_n$ but only the subconcept $u_3 = \textbf{animal with breasts hairy}:s_3,...,s_n$, but for the transformation of this concept in a natural language we permit any correct permutation of its attributes 'hairy' and 'with breasts', by conserving the bijection between the canonical subconcepts and predicate atoms,\\
    $p_m^n(y_1,y_2,,...,y_n) ~~~~~\Leftrightarrow~~~~~ \textbf{animal}:s_1,...,s_n \in D_n$, for a predicate-concept,\\
   $p_m^n(g(y_1),y_2,,...,y_n) ~~~~~\Leftrightarrow~~~~~ \textbf{animal with breasts}:s_2,...,s_n\in D_{n-2}$,\\
   $p_m^n(y_1,g(y_2),,...,y_n) ~~~~~\Leftrightarrow~~~~~ \textbf{animal    hairy}:s_2,...,s_n\in D_{n-2}$,\\
   $p_m^n(g(y_1),g(y_2),,...,y_n) ~~~~~\Leftrightarrow~~~~~ \textbf{animal with breasts hairy}:s_3,...,s_n\in D_{n-3}$,\\
    etc...

 Thus, we have the following requirement for the correspondence between the concepts an predicates in IFOL:
 \begin{coro} \label{coro:concept-pred}
 \textsc{C-structure completeness}: The correspondence between the concepts and predicates in the IFOL has to satisfy the following requirement:\\
% 1. Any unary concept in $D_1$, which is not a derived subconcept of another n-ary concept (with $n\geq 2$), must be a predicate-concept of some unary predicate in IFOL,\\
  Any k-ary concept $u \in D_k$, $k\geq 1$, has to be a predicate-concept or a canonical subconcept of some predicate-concept.\\
  In such a C-structure, we preserve the bijective correspondence between the concepts and predicate atoms in IFOL, by using only canonical composition of the attributes in the subconcept's names.
 \end{coro}
 \textbf{Proof}:  Easy to verify.
 \\$\square$\\
 By the assignments $g$ that assign the values to the free variables in $p_m^n(y_1,...,x_n)$, such that at least one of them remains free, we obtain the complete tree where the nodes are the subconcepts of the predicate-concept and arcs between nodes are the IS-A relationship $'\sqsubseteq'$: the origin of this tree is the predicate-concept, while the leafs of the tree are its \emph{unary} susbconcepts.
 So, intuitively we require that if we use any concept $u$, different from the leafs of such a predicate-concept tree,
   %$\textbf{animal}:s_1,...,s_n$ in $ D_{n}$
 \emph{as a sort} of another concepts in $\D$, then the extension of such a sort $u$ has to contain the \emph{names} of all its subconcepts $u_i$ ( with $u_i \sqsubseteq u$). Hence, in our example, if we want to use the n-ary predicate-concept $\textbf{animal}:s_1,...,s_n$ in $ D_{n}$ as a sort, the subconcept's \emph{name} \textbf{animal with brests} will be an element of this sort of animals: in this way, the n-ary concept $\textbf{animal}:s_1,...,s_n$ when used as \emph{a sort} will have no more an n-ary relation but an unary relation as its sort-extension.\footnote{While the standard extension of the n-ary concept $\textbf{animal}:s_1,...,s_n$, corresponding to the predicate atom $p_m^n(y_1,y_2,,...,y_n)$, is a n-ary relation\\
   $h(\textbf{animal}:s_1,...,s_n) = \{(g(y_1),...g(y_n))\mid g\in \D^\V$, \\such that $I_T(p_m^n(g(y_1),...g(y_n))) = t\}$.}

 The IS-A relationship, based on Definition \ref{def:var-sort}, provides a hierarchy over the concepts and provides the basis for the “inheritance of properties”: when a concept is more specific than some other concept, it inherits the properties of the more general one.
 We restrict this inheritance property only to the concept in relation with the same predicate, that is, inside the IS-A tree bellow a top predicate-concept. In this way, if the robot learned that some concept has also another attribute (a sort), by the inheritance of properties this attribute is added to each subconcept of the rot predicate-concept, and a new variable of the sort of this new attribute is added also to the associate predicate.  This is the typical setting of the so-called (monotonic) \emph{inheritance networks} (see \cite{Brac79}).
 For example, if the \textbf{animal} has an age (for example the sort $s_2$ above), then the \textbf{animal with breasts}  has an age attribute too (while for the predicate-concept with the name \textbf{cat}, such that $\textbf{cat}\sqsubseteq \textbf{animal}$ it is not necessary requirement): between two different predicate-concepts (that define the predicates in IFOL) in an IS-A relationship, we have no necessarily the inheritance of properties.

 With this relaxation, we do not impose to have the predicates with enormous number of attributes  hereditable  from all concepts in the ontology subtree with the root (top concept) corresponding to its predicate-concept.

However, for the subconcept $\textbf{animal with breasts}:s_2,...,s_n$ in $D_{n-1}$ we can generate the same concept but with different name $\textbf{mamal}:s_2,...,s_n$ in $D_{n-1}$, and to use it in the tree of the subconcepts of the predicate-concept $\textbf{animal}:s_1,s_2,...,s_n$ in $D_{n}$ as an equivalent substitution of the \textbf{animal with breasts} by preserving the inheritance property described above.\\
 \textbf{Remark}:
 So, the IS-A classification between predicate-concepts can be used independently from the sets of the currently defined (learned) set of the attributes (sorts) of these different concepts. Automatic inheritance of the attributes remains only between the subconcepts of a given predicate-concept (it is necessary because all of them are in relationship with \emph{the same} predicate).\\
 Moreover, from the fact that any, also partial, assignment to free variables of the predicate generates a particular subconcept of the predicate-concept, it is important to limit such assignments only to really significant values for such variables and not to all values in the domain $\D$.

 For the built-in sorts like \textbf{reals}, \textbf{integers},  \textbf{data}, etc.., it is important tu use the IS-A relationship, like $\textbf{natural numbers}\sqsubseteq \textbf{integers}\sqsubseteq \textbf{rationals}\sqsubseteq \textbf{reals}$, during mathematical computations and the necessity to use not only this statically defined sorts but also the dynamic sorts, as it will be explained in next.
  \\$\square$\\
 Note that we have the family of the vedrb.based predicate-concept by names, for example, \textbf{to work}, \textbf{to go}, \textbf{to know}, \textbf{to like}, \textbf{to be}, etc.., with this particular propriety that they are not the unary concepts, but the relation concepts in $D_k$ for any $k\geq 2$, such that the first sort (argument) of each verb-based concept is the sort $s_1 = \textbf{verb form}$ which is built-in unary concept $u = \textbf{verb form}:\textbf{verb form} \in D_1$, which fixed extension $h(\textbf{verb form})$ contains as extension all possible forms of the verbs used in grammar, like the values 'present', 'past', 'future', 'gerund', etc..., as we used in the Examples 2 and 3 in previous Section 3, and in next example.

 %--------------------------------------------------------------
We denote by $\top_S\in D_1$ the most general predicate-concept  $\textbf{everything}:\textbf{everything}\in D_1$ of the sort \textbf{everything} containing the set of all finite natural language expressions %for which $h(\top)  = \D$,
and by  $\bot_S \in D_1$ the bottom predicate-concept $\textbf{empty set}:\textbf{empty set}$  of the sort \textbf{empty set} with empty extension $\|\textbf{empty set}\| = h(\bot_S) = \emptyset$, so that for each k-ary concept $u \in D_k$ for $k\geq 1$,
$$\bot_S \sqsubseteq u \sqsubseteq \top_S$$
So, from the fact that for each natural language the lexicon is finite, the set of concepts is finite as well, and hence the IS-A ontology of the PRP concepts is a complete lattice of the concepts.  In such IS-A ontology hierarchy (lattice) where the nodes (denoted in bold letters) are the concepts graphically connected by the IS-A arcs. So, we assume that the set of all sorts $S$ for the robots is finite and enough big for them. Otherwise, we can use also infinite set of sorts, but we require that in this case it has only \emph{Noetherian sort hierarchy}\footnote{
For example, Java generic classes (as introduced in version 5 of the programming language) inherently lead to non-Noetherian type systems.}
 $(S,\sqsubseteq)$, such that every infinite descending chain eventually becomes stationary, that is, for every chain of sorts in $S$,  $s_0 \sqsupseteq s_1\sqsupseteq...$, there exists a natural number $~n_0$ such that $s_n= s_{n_0}$ for all $n \geq n_0$.
\\
\textbf{Remark}:
 Note that in IFOL we do not introduce the binary predicate for the IS-A relationship '$\sqsubseteq$', because it there exists only for the concepts and we only extend the intensional algebra $\A_{int}$ in \cite{Majk22} with this relationship and restrict the set of extensionalization functions $h \in \E$ to the sorted-subset $\E_s \subset \E$ and hence to the subset of many-sorted Tarski's interpretations in which, for a fixed intensional interpretation $I$, the many-sorted Tarski's interpretations $I_T^* =h\circ I$ for some $h\in \E_s$.
  \\
  The sorted-subset extensionalization functions $h \in \E_s$ has to satisfy the set-inclusion of the sort's extensions $\|s_1\|\subseteq\| s_2\|$ if robot learned that $s_1 \sqsubseteq s_2$, and we do not use '$\sqsubseteq$' for the axiomatization of Many-sorted IFOL.
  \\$\square$\\
When we evaluate a term using some interpretation, we get an element of the
domain.  For example \cite{Gies05}, let us assume two sorts $\mathbb{Z}$, signifying the integers, and $\mathbb{Q}$, signifying the rationals. We want $\mathbb{Z}$ to be a subsort of $\mathbb{Q}$. The numbers $1,2,3,...$ (particulars  in $D_{-1}$) are
constants of static type $\mathbb{Z}$. The operation of division $div: \mathbb{Q}\times \mathbb{Q} \rightarrow \mathbb{Q}$ produces a term
of static sort $\mathbb{Q}$ from two terms of static sort $\mathbb{Q}$. Since $\mathbb{Z}$ is a subsort of $\mathbb{Q}$, $\mathbb{Z}\sqsubseteq\mathbb{Q}$
we can use two as argument for division function. The static sort of the term $ t = div(2, 2)$ is $\mathbb{Q}$, since that is the return sort of functional symbol $f^2_i = div$. The domain of the standard model
consists of the set of rational numbers, where the integers have \emph{dynamic sort} $\mathbb{Z}$, and all other elements have dynamic sort $\mathbb{Q}$. In the standard model, the value of $div(2, 2)$ is the domain element $1$. The \emph{dynamic} sort of the value of $div(2, 2)$ is therefore $\mathbb{Z}$, which is all right since $\mathbb{Z}$ is a subsort of $\mathbb{Q}$.

The semantics of a term is going to be a value in the domain $\D$. Each domain
element $u\in \D$ has a dynamic sort $\delta(u) \in S_D$. While each domain element has a unique dynamic sort, it may still occur as the value of terms of different static sorts. Our semantic definitions will be such, that the dynamic sort of the value of a term is guaranteed to be a subsort of the static sort of the term, that is,  $\delta(u) \sqsubseteq \digamma(u)$.
%We denote the set of valid domain elements for a certain static type by

Having in mind this dynamic property, we can define the sets of sorted terms of a given static sort $s\in S_D$, and the many-sorted formulae of Many-sorted IFOL:
\begin{definition} \label{def:M-Sort terms}
We inductively define the system of sets:
\begin{enumerate}
  \item The sets $\{T_s\}_{s \in S_D}$ of terms of
static sort $s\in S_D$ to be the least system of sets such that\\
- $x \in T_s$ for any variable $x:s$, that is with  $\digamma(x) = s$,\\
- $f^k_i(t_1,...,t_k) \in T_s$ for any k-ary functional symbol $f^k_i\in F$ such that $\digamma(f^k_i) = \digamma(t_1)\times...\times\digamma(t_k)\times s$, and terms $t_i \in T_{s_i}$ with $s_i \sqsubseteq \digamma(t_i)$, for $1\leq i\leq k$.
  \item The set of formulae $\F$ to be the least set
such that\\
- $p^k_i(t_1,...,t_k) \in \F$ for any k-ary  predicate  symbol $p^k_i\in P\backslash F$ such that $\digamma(p^k_i) = \digamma(t_1)\times...\times\digamma(t_k)$, and terms $t_i \in T_{s_i}$ with $s_i \sqsubseteq \digamma(t_i)$, for $1\leq i\leq k$,\\
- for the rest it is equal as for unsorted case.
\end{enumerate}
\end{definition}
Thus, our semantics is defined in such a way that the sort of the value of a term will be a subsort of, or equal to the static sort of the term. When we mean this notion of sort, we will talk about the \emph{dynamic sort} of a domain element.
\begin{definition} \label{def:DynSort} \textsc{Many-sorted assignments}:
In the presence of dynamic sorts as well, the set of valid domain elements $D_s$ for a certain static sort $s \in S_D$ is possibly greater that that given by Definition \ref{def:Static Sorts}, that is,
\begin{equation} \label{eq:DynSort}
D_s = \{u \in \D \mid \delta(u) \sqsubseteq s\}~~~~\supseteq ~~~~\|s\|
\end{equation}
Consequently, we can define the set of many-sorted assignments $G_S \subset \D^\V$ restricted w.r.t the unsorted assignment in $\D^\V$ , such that for any variable $x_i \in \V$, and many-sorted assignment $g \in G_S$,
%\begin{equation} \label{eq:MSassign}
$g(x_i) \in D_s$, for the variable's sort $s = \digamma(x_i)$.\\
We then extend it inductively (analogoulsy to that  for unsorted FOL) into a many-sorted assignment $g^*:\T \rightarrow \D$, where $\T$ denotes the set of all terms, for a given Tarski's interepretation $I_T$ of the IFOL , by:
\begin{enumerate}
  \item $g^*(t_i) = g(x) \in D_s$ if the term $t_i$ is a variable $x \in \V$ of sort $s = \digamma(x)$.
  \item $g^*(t_i) = I_T(c) \in D_s$ if the term $t_i$ is a constant $c \in F$ of sort $s = \digamma(I(c))$ of particular $I(c) \in D_{-1}$.
  \item If a term $t_i$ is $f_i^k(t_1,...,t_k)\in T_s$, where $f_i^k \in F$ is a
k-ary functional symbol such that its sort is $\digamma(f^k_i) = s_1\times...\times s_k\times s$ and $t_i\in T_{s_i}$ are terms of sort $s_i \sqsubseteq \digamma(t_i)$ for $1\leq i \leq k$, then
$g^*(f_i^k(t_1,...,t_k)) = I_T(f_i^k)(g^*(t_1),...,g^*(t_k))$  where\\
$I_T(f_i^k): D_{s_1} \times...\times D_{s_k} \rightarrow D_s$\\
 is a function or, equivalently, in the graph-interpretation of the function,\\
 $g^*(f_i^k(t_1,...,t_k)) = u \in D_s$  such that $(g^*(t_1),...,g^*(t_k),u) \in I_T(f_i^k)$\\
 with $I_T(f_i^k) \subseteq D_{s_1} \times...\times D_{s_k}\times D_s$, where in IFOL also functional symbols are considered as predicates.
 %(as in (\ref{eq:modelFOLk})).
\end{enumerate}
The extension of Tarski's interpretation $I_T$ to its generalized interpretation $I_T^*$ is identical to that provided in \cite{Majk22}, by considering now that $g\in \D^\V$ are only the many-sorted assignments.
\end{definition}
The Tarski's constraints in \cite{Majk22} remain identical by  using the many-sorted assignments $g$ (instead of unsorted).\\
With this, a model $\M = (\D,I_T)$ and satissfaction relationship is identical to that inunsorted FOL by considering only the many-sorted assignments $g$.
So, with this we obtain  $I_T^* = h\circ I$ for a fixed intensional interpretation $I$, with the many-sorted restriction of the extensionalization functions $h\in \E_S \subset \E$, such that for any n-ary concept $\textbf{phrase}:s_1,s_2,...,s_n$ in $D_{n}$, for $n\geq 1$, it must hold that

$h(\textbf{phrase}:s_1,s_2,...,s_n) \subset D_{s_1} \times...\times D_{s_n}$\\

 We represented the Tarski's FOL interpretation by new single mapping $I_T^*:\L\rightarrow \mathfrak{R}$, as explained previously. But also intensional semantics is given by $h\circ I:\L\rightarrow\mathfrak{R}$ where $\circ$ is a
composition of functions.
So, if there is a modal Kripke semantics with a set of Kripke's possible
worlds  (thus, an intensional semantics) for FOL, \emph{equivalent} to the standard FOL semantics given by the Tarski's interpretation $I_T$, then we obtain the correspondence $I_T^* = h \circ I$, expressed by the commutative diagram for \emph{unsorted} Intensional FOL \cite{Majk22},  such that for any  $\phi \in \L$, $~h(I(\phi)) = I_T^*(\phi)$.

It has been demonstrated by Theorem 1 in \cite{Majk22}, Section 1.2.1,  for the unsorted IFOL.
In the many-sorted IFOL aqnalogouasly we have that for  each Tarski's interpretation $I_T \in \mathfrak{I}_T(\Gamma)\}$, such that its generalization $I_T^* = \hbar \circ I$ for actual Montague's world $w_0$ (corresponding to the extensionalization function $\hbar =\F(w_0)   \in \E_S$) of a given set of assumptions $\Gamma$, there is a unique Kripke's  interpretation $~I_K: P \rightarrow {\bigcup}_{n \in \N} \textbf{2}^{\D^n}$, and vice versa, such that for any $p_i^k \in
 P\backslash F$ of sort $\digamma(p_i^k) = s_1\times...\times s_k$, and $f_i^k \in F$ of sort $\digamma(f_i^k) = s_1\times...\times s_k\times s$, $k\geq 1$,  with $(d_1,...,d_k) \in D_{s_1} \times...\times D_{s_k}$ and $(d_1,...,d_k,d_{k+1}) \in D_{s_1} \times...\times D_{s_k}\times D_s$, and any intrinsic world (many-sorted assignment) $g\in \D^\V$, it holds that:

 $I_K(p_i^k)(d_1,...,d_k) = t~~$ iff $~~(d_1,...,d_k) \in I_T(p_i^k)~~$ and

 $I_K(f_i^k)(d_1,...,d_k,d_{k+1}) = t~~$ iff $~~ d_{k+1} =
 I_T(f_i^k)(d_1,...,d_k)$.\\
 So, we van define the set $\mathfrak{I}_K(\Gamma) = \{I_K$ defined above from $I_T~|~I_T \in \mathfrak{I}_T(\Gamma)\}$  with the bijection  between the sets of Tarski's and Kripke's interpretations $$~\flat:\mathfrak{I}_T(\Gamma) \simeq
 \mathfrak{I}_K(\Gamma)$$ so that for any Tarski's interpretation we
 have its equivalent Kripke's interpretation $I_K = \flat(I_T)$.

Thus, for the \emph{minimal} unsorted IFOL with the syntax given by the algebra $\A_{FOL}$ in \cite{Majk22}, and with the abstracted terms as well, we have the following Kripke modal semantics (where the semantics of the existential quantifiers $\exists_i$ in $\A_{FOL}$ are expressed by the existential \emph{modal operators} $~\lozenge_i$ with the accessibility relations $\R_i \subset \D^{\V}\times \D^{\V}$ defined in the following definition):
\begin{definition} \label{def:IntensionalFOL} \textsc{Many-Sorted Intensional FOL (FOL$_{\I}(\Gamma)$) Semantics:}

  Let $\M = (\mathbb{W}, \{{\R}_i \}, \D, I_K)$ be a Kripke's model of a predicate
multi-modal logic, obtained from FOL$(\Gamma)$ with a set of its Tarski's interpretations $\mathfrak{I}_T(\Gamma)$ which are models of $\Gamma$, with the set of  possible worlds $\mathbb{W}$ is the set of \emph{many-sorted} assignments $G_S\subset \D^\V$, and the set of existential modal "low-level" operators $\diamondsuit_i$  for each existential modal operator $\exists_i$ in $\A_{FOL}$ (corresponding to FOL quantifier $(\exists x),~x\in \V$) with the accessibility relation $~{\R}_i = \{(g_1,g_2) \in G_S\times G_S~ |$ for all $y \in \V \backslash \{x\}(g_1(y) = g_2(y))\}$.
 %as it was shown in Definition \ref{def:FOL-KripSem}.

 The mapping $~I_K: P \rightarrow {\bigcup}_{n \in \N} \textbf{2}^{\D^n}$ is a many-sorted  Kripke's interpretation such that, for a given Tarski's  many-sorting interpretations of FOL$(\Gamma))$, $I_T^* \in \mathfrak{I}_T(\Gamma)$, and a predicate $p_i^k \in P\backslash F$, of sort $\digamma(p_i^k) = s_1\times...\times s_k$, $k\geq 1$,  with $(d_1,...,d_k) \in D_{s_1} \times...\times D_{s_k} \subset \D^k$, we have that $I_K(p_i^k)(d_1,...,d_k) = ~I_T^*(p_i^k(d_1,...,d_k))$.
% , that is, from Theorem \ref{Th:FOL-KripSem}, $I_K = ~\flat(I_T^*)$ .

 Then we define the Kripke's interpretation of the Intensional logic
 FOL$_{\I}(\Gamma)$ by $\M_{FOL_{\I}(\Gamma)} = \widehat{\M} =_{def} (\widehat{\mathbb{W}}, \{{\R}_i \}, \D, \widehat{I}_K)$
 with  a set of (generalized) possible worlds $\widehat{\mathbb{W}} = \mathfrak{I}_T(\Gamma) \times G_S$, where $\W_e = \mathfrak{I}_T(\Gamma)$ is the set of explicit possible  worlds (the set of Tarski's many-sorting interpretations of FOL$(\Gamma))$, non empty domain $\D$, and  the mapping  $~~\widehat{I}_K:\W_e \times P \rightarrow {\bigcup}_{n \in \N}\textbf{2}^{\D^n}$ such that for each $I_T^* \in \W_e$ and $p_i^k \in P\backslash F$,  of sort $\digamma(p_i^k) = s_1\times...\times s_k$, $k\geq 1$,  with $(d_1,...,d_k) \in D_{s_1} \times...\times D_{s_k} \subset \D^k$, we have  $~\widehat{I}_K(I_T^*,p_i^k)(d_1,...,d_k) = I_K(p_i^k)(d_1,...,d_k)$ where $I_K = \flat(I_T^*)$.\\
 For each world $\textbf{w} = (I_K,g) \in \widehat{\mathbb{W}}$, where $g$ is a many-sorted assignment to variables in $\V$, we have that
 \begin{enumerate}
 \item For each atom $p_i^k(t_1,...,t_k)$ of with sort of its predicate $\digamma(p_i^k) = s_1\times...\times s_k$, $k\geq 1$, and terms $t_i \in T_{s_i}$ for $1\leq i \leq k$,\\
$~~{\widehat{\M}} \models_{I_K,g}~p_i^k(t_1,...,t_k)~~~$ iff $~~~\widehat{I}_K(I_T,p_i^k)(g^*(t_1),...,g^*(t_k)) = t$,
 \item $~~{\widehat{\M}} \models_{I_K,g}~ \varphi \wedge \phi~~~$ iff
$~~~{\widehat{\M}} \models_{I_k,g}~ \varphi~$ and $~{\widehat{\M}}
\models_{I_K,g}~ \phi~$,
 \item $~~{\widehat{\M}} \models_{I_K,g}~ \neg \varphi ~~~$ iff $~~~$ not ${\widehat{\M}} \models_{I_K,g}~ \varphi~$,
 \item $~~{\widehat{\M}} \models_{I_K,g}~\lozenge_i \varphi~~~$ iff $~~~$
exists $g'\in G_S$ such that $(g,g') \in {\R}_i $ and
${\widehat{\M}} \models_{I_K, g'}~ \varphi$.
 \end{enumerate}
  \end{definition}
Let us show that this unique intensional Kripke model $\M_{FOL_{\I}(\Gamma)} = (\widehat{\mathbb{W}}, \{{\R}_i \}, \D, \widehat{I}_K)$  models the Tarskian \emph{logical consequence} of the  First-order logic with a set of assumption in $\Gamma$, so that the added intensionality preserves the Tarskian semantics of the FOL.
\begin{propo}  \label{prop:IntensionalFOL} \textsc{Montague's intension and Tarski's interpretations}:

Let $\M_{FOL_{\I}(\Gamma)} = (\widehat{\mathbb{W}}, \{{\R}_i \}, \D, \widehat{I}_K)$ be the  unique \verb"intensional" Kripke model  of the  First-order logic with a set of assumptions in $\Gamma$, as specified in Definition \ref{def:IntensionalFOL}.

So, a formula $\phi$  is \emph{a logical consequence} of $\Gamma$ in the Tarskian semantics for the FOL, that is, $~~\Gamma\Vdash \phi,~~$ iff $~~\phi$ is true in this Kripke intensional model   $\M_{FOL_{\I}(\Gamma)}$, i,e, iff $~\|\phi\| = \widehat{\mathbb{W}} = \W_e\times G_S$ with the set of Tarski's interpretations which are models of $\Gamma$ (explicit worlds) $\W_e = \mathfrak{I}_T(\Gamma)$.

 If $I_n:\L_{op} \rightarrow \mathfrak{R}^{\W_e}$ is the mapping given in (\ref{eq>Montague}) then, for any (virtual) predicate $\phi(x_1,...,x_k) $, the mapping $I_n(\phi(x_1,...,x_k)):\W_e\rightarrow \mathfrak{R}$  represents the Montague's meaning (intension) of this logic formula, such that for any Tarski's interpretation (an explicit possible world) $~I_T^* = w \in \W_e = \pi_1(\mathbb{W})$,
\begin{equation} \label{eq:MotagTarski}
I_n(\phi(x_1,...,x_k))(w) = w(\phi(x_1,...,x_k))= I_T^*(\phi(x_1,...,x_k))
\end{equation}
 \end{propo}
\textbf{Proof:} Let us show that for any first-order formula $\phi$ and a many-sorted assignment $g\in G_S \subset \D^\V$
it holds that, $~~\M_{FOL_{\I}(\Gamma)} \models_{w,g}~ \phi ~~$ iff
$~~w(\phi/g) = t$, where $w$ is the  Tarski's
interpretation $w = I_T^* \in \W_e = \mathfrak{I}_T(\Gamma)$. Let us demonstrate it by the structural induction on the length of logic formulae. For any
 atom $p_i^k(t_1,...,t_k)$ of with sort of its predicate $\digamma(p_i^k) = s_1\times...\times s_k$, $k\geq 1$, and terms $t_i \in T_{s_i}$ for $1\leq i \leq k$,
 we have from Definition \ref{def:IntensionalFOL} that

$\M_{FOL_{\I}(\Gamma)} \models_{I_T^*,g}~p_i^k(t_1,...,t_k)~~~$
iff $~I_K(I_T^*,p_i^k)(g^*(t_1),...,g^*(t_k)) = t~~~$

 iff $~~~I_T^*(p_i^k(t_1,...,t_k)/g) = t$. \\
 Let us suppose that such a property holds for every formula $\phi$ with less than $n$ logic connectives of the FOL, and let us show that it holds also for any
formula with $n$ logic connectives. There are the following cases:
\begin{enumerate}
  \item The case when $\phi = \neg \psi$ where $\psi$ has $n-1$ logic
connectives. Then $~~\M_{FOL_{\I}(\Gamma)} \models_{I_T^*,g}~ \phi~~~$
iff $~~~\M_{FOL_{\I}(\Gamma)} \models_{I_T^*,g}~ \neg \psi ~~~$ iff
$~~~$ not $\M_{FOL_{\I}(\Gamma)} \models_{I_T^*,g}~ \psi ~~~$ iff (by
inductive hypothesis) $~~~$ not $I_T^*(\psi/g) = t~~~$ iff  $~~~I_T^*(\neg\psi/g) = t~~~$ iff  $~~~I_T^*(\phi/g) = t$.
  \item The case when $\phi =  \psi_1 \wedge \psi_2$, where both $\psi_1,
\psi_2$ have less than $n$ logic connectives, is analogous to the case 1.
  \item The case when $\phi = (\exists x) \psi$ with the free variable $x:s$ of sort $s$,  where $\psi$ has $n-1$
logic connectives.
%It is enough to consider the case when $x$ is a free variable in $\psi$.
Then $~~\M_{FOL_{\I}(\Gamma)}
\models_{I_T^*,g}~ \phi~~~$ iff $~~~\M_{FOL_{\I}(\Gamma)}
\models_{I_T^*,g}~ (\exists x) \psi ~~~$ iff $~~~$ exists an element of sorts $s$, $u \in D_s$,
such that $~~\M_{FOL_{\I}(\Gamma)} \models_{I_T^*,g}~ \psi[x/u]~~~$
iff (by inductive hypothesis)  $~~~$ exists $u \in D_s$
such that $~~I_T^*(\psi[x/u]/g) = t~~~$  iff $~~~I_T^*((\exists x)\psi/g) = t ~~~$ iff $~~I_T^*((\phi/g)) = t$.
\end{enumerate}
It is easy to verify that the intension of predicates in the FOL$_{\I}(\Gamma)$ defined above can be expressed by the mapping $I_n$ such that for any $p^k_i \in P\backslash F$ and $w =I_T^* \in \W_e$, $I_n(p_i^k(t_1,...,t_k))(w) = w(p_i^k(t_1,...,t_k))=I_T(p_i^k(t_1,...,t_k))$, and, more
general, for any virtual predicate (where all its terms $t_i$ are just free variables $x_i$ of the sorts $\digamma(x_i) = s_i$), $\phi(x_1,...,x_k)$, for a many-sorted assignment $g \in G_S \subset \D^\V$,

 $I_n(\phi(x_1,...,x_k))(w) =\|\phi(x_1,...,x_k)\|$

 $= \{ (g(x_1),...,g(x_k)) \in D_{s_1}\times...\times D_{s_k}~| ~g \in G_S$ and  ${\M}_{FOL_{\I}}
\models_{w,g}~ \phi(x_1,...,x_k) \}$

$= \{ (g(x_1),...,g(x_k))  ~| ~g \in G_S$ and $w(\phi(x_1,...,x_k)/g) = t \}$

$= \{ (g(x_1),...,g(x_k))  ~| ~g \in G_S$ and
$(g(x_1),...,g(x_k)) \in w(\phi(x_1,...,x_k)) \}$

$= w(\phi(x_1,...,x_k)) = I_T^*(\phi(x_1,...,x_k))$,\\
where $w$ is the  Tarski's interpretation $w =I_T^*
\in \W_e = \mathfrak{I}_T(\Gamma)$.\\ So,
$I_n(\phi(x_1,...,x_k)):\W \rightarrow \mathfrak{R}$ is the Montague's meaning (i.e., the intension) of the (virtual) predicate $\phi(x_1,...,x_k)$.
\\$\square$\\
The main difference between Tarskian semantics and this intensional
semantics is that this unique intensional Kripke model $\M_{FOL_{\I}(\Gamma)}$ encapsulates the set of all Tarski models of the First-order logic with a (possibly empty) set of assumptions $\Gamma$.
\section{Conclusions and future work}
The Intensional FOL (IFOL) generalizes both approaches (Description Logic and Conceptual Graphs as continuation is a subset of the IFOL where the existential quantifiers have a modal semantics), and by using the abstract terms (reification of the sentences into another sentences) is the most adequate formal language to represent the natural languages: its logical syntax based on the predicates support the Tarski's semantics as well, while human cognitive intuitions and in their features are supported by its intensional concepts, its structure and intensional algebra.

 The fact that FOL is theoretically undecidable logic reasoning system, differently from the simpler conceptual-based versions of  Description Logic is not the problem for using IFOL as formal cognitive and reasoning system for AGI (Strong AI) robots, because we need that they support also uncomplete (unknown facts) and mutually-contradictory information, like the humans do, and hence for this necessity we can use the 4-valued (Belnap's bilattice) truth system where both with truth and falsity we have "unknown" and "possible" logic values for the logic sentences as provided in \cite{Majk22}.  In this many-valued version of the IFOL the value "unknown" would be assigned to each sentence for which the deduction process overcome some interval of time (thus, undecidable with high probability), lake the humans do.

% ---- Bibliography ----
%

\end{document}